# A Soft Robotic Exosuit For Knee Extension Using Hyper-Bending Actuators

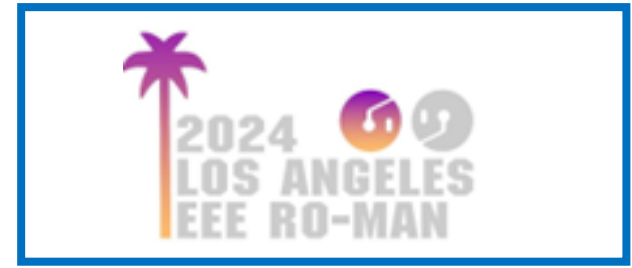

Tuo Liu[1], Jonathan Realmuto[2]

[1] tliu189@ucr.edu
[2] jonathan.realmuto@ucr.edu
Bionic Systems Lab, Department of Mechanical Engineering, University of California, Riverside

Movement disorders impact muscle strength and mobility, and despite therapeutic efforts, many people with movement disorders have challenges functioning independently. Soft wearable robots, or exosuits, offer a promising solution for continuous daily support, however, commercially viable devices are not widely available. Here, we introduce a design framework for lower limb exosuits centered on a soft pneumatically driven fabric-based actuator. Our design consists of a novel multi-material textile sleeve that incorporates braided mesh and knit-elastic materials to realize hyper-bending actuators. The actuators incorporate 3D-printed self-sealing end caps that are attached to a semi-rigid human-robot interface to secure them to the body. We will demonstrate the effectiveness of our exosuit in generating enough force to assist during sit-to-stand transitions.

## INTRODUCTION

Soft wearable robots, or *exosuits*, are a promising solution for delivering constant physical support to the mobility impaired [3] because they can be made lightweight, compliant, and safe. However major challenges include developing soft actuators that can deliver adequate force while maintaining a small volumetric form factor and attaching these actuators to the body in a comfortable way with efficient force transmission.

We have designed a new hyper-bending fabric actuator by leveraging compliance in two independent directions across the surface of an inflatable. Stretch along the longitudinal direction of the knit-elastic results in elongation while radial expansion of the braided mesh produces shortening. The hyper-bending actuator, shown in Fig. 1, combines the two behaviors into a cylindrical unit and produces more bending when compared with knit-elastic combined with inelastic fabric, as previously developed [2-4]. To complete the exosuit, the hyper-bending actuators are engineered to be easily mounted to custom pants to orient them parallel to the knee so that the bending motion provides extension torques. Our design is flexible and modular and can be adapted to different assistive applications.

## DESIGN AND FABRICATION FRAMEWORK

The mesh braided material used to construct the hyper-bending actuator is the same material from the classic McKibben artificial muscle [1], and the knit-elastic material, commonly used as waistbands in garments, has been extensively used by our research group for developing bending and twisting actuators [2-4]. These two materials sewn together form the actuator sleeve, which is the functional part of the actuator and determines the resulting motion. In general the sleeve can be constructed with different materials to program different motions. A silicone bladder with air inlets in placed in the sleeve and 3D-printed end-cap are used to mate the sleeve and bladder, as shown in Figures 2 and 3.

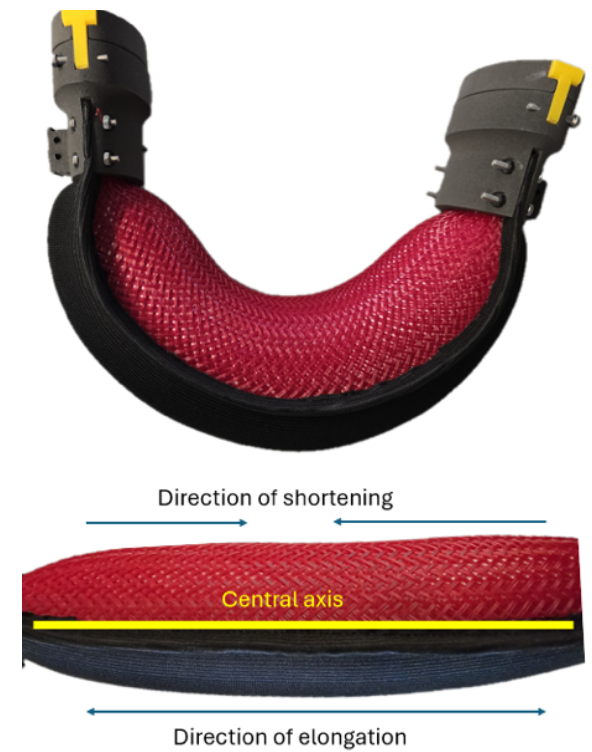

**Figure 1.** Working principle of the hyper-bending actuator. Top: Inflated actuator; Bottom: principle of deformation with knit-elastic (bottom fabric) elongating and braided mesh (top fabric) side shortening.

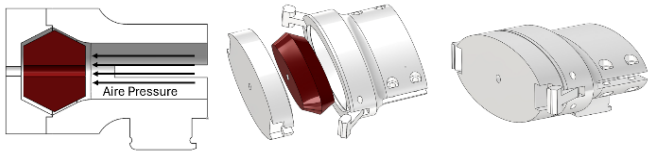

**Figure 2.** Renders of the self-sealing 3D printed end-caps. Left: cross section view of end-cap; Center: Top and bottom end-cap with air seal collar; Right: End-cap with locking hinge closed.

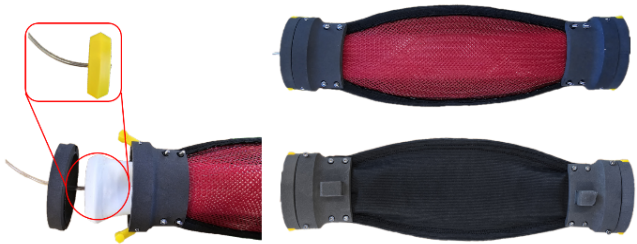

**Figure 3.** Details of the actuator assembly. The air inlet uses a custom inner collar with the end-cap top and bottom piece compressing the collar and bladder together. The two sleeve layers sewn together with the internal bladder are held together by the 3D printed end-caps.

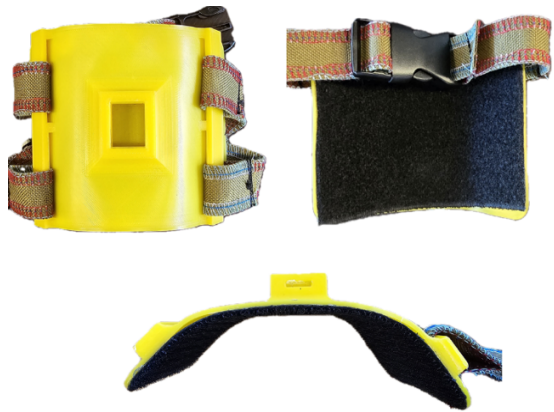

**Figure 4.** The 3D printed human-robot interface for securing the actuator to the body. The system includes hook and loop fabric that secures to corresponding hook/loops sewn into the custom pants.

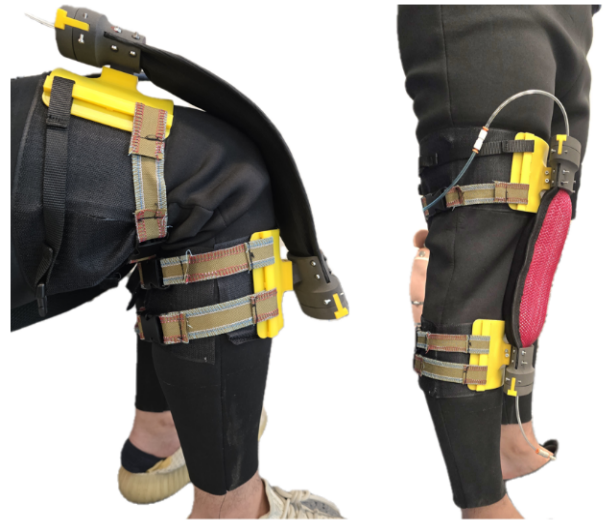

**Figure 5.** Our proposed exosuit. Left: Deflated state with user sitting. Right: Inflated state with user standing. Note that the shape of the actuator appears slender because the human is loading the device.

To attach the actuators to the body, we designed a human-robot mounting interface, as shown in Figure 4. This 3D-printed interface includes a mounting feature for the actuators mounting and it conforms to the shape of the human limb. The backside is lined with adhesive hook/loop fabric, allowing it to securely attach to custom hook/loop fabric on the pants with additionally straps further securing it to the limb. The design ensures a large contact area for distributing the force generated by the actuator to minimizing discomfort.

The exosuit system consists of three main components: the hyper-bending actuator, the human-robot mounting interface, and custom-made neoprene pants, and the complete system is shown in Figure 5. The pants include a hook/loop interface that corresponds to the human-robot hook/loop. This allows for easy donning of the actuators and provide a level of modularity. fabric is glued and sewn onto the pants for added strength. The mounting interface is attached to the loop fabric and secured with straps.

Our preliminary testing has shown the exosuit can provide and transmit substantial forces to the body. For our demonstration we will showcase individual components including the actuators and interfaces, and we will show the device assisting a user during sit-to-stand transitions.

## CONCLUSIONS

This work presents a design and fabrication framework for assistive exosuits for the lower limbs. Our main innovation is a hyper-bending actuator that can effectively transmit high forces in a low volumetric workspace. Compared to existing solutions, our design reduces the effort required to don a wearable robot onto the body. We aim to include more degrees of freedom, including hip extension in future efforts.

## ACKNOWLEDGMENT

This research was supported by *A Foundation Building Strength*.